\documentclass{amia}
\usepackage{graphicx}
\usepackage[labelfont=bf]{caption}
\usepackage[superscript,nomove]{cite}
\usepackage{color}

\usepackage{amssymb}
\usepackage{changepage}
\usepackage{amsmath,amssymb}
\usepackage{caption}
\usepackage{subcaption}
\usepackage{xcolor} 
\usepackage[framemethod=tikz]{mdframed}
\usepackage{mathtools}

\usepackage{url}
\usepackage{makecell}
\usepackage{multirow}
\usepackage[section]{placeins}
\usepackage[linesnumbered,lined,boxed,commentsnumbered]{algorithm2e}

\usepackage{xcolor}
\definecolor{cb_orange}{rgb}{1.0,0.51,0.0}
\definecolor{cb_blue}{rgb}{0.22,0.49,0.72}
\definecolor{cb_green}{rgb}{0.3,0.67,0.29}
\definecolor{cb_red}{rgb}{0.89,0.1,0.11}
\definecolor{cb_purple}{rgb}{0.6, 0.31, 0.64}
\definecolor{cadetgrey}{rgb}{0.57, 0.64, 0.69}

\begin{document}

\title{Deep EHR Spotlight: a Framework and Mechanism to Highlight Events in Electronic Health Records for Explainable Predictions}

\author{Thanh Nguyen-Duc, MSc$^{1,2}$; Natasha Mulligan, MSc$^{1}$; Gurdeep S. Mannu, MBBS, MRCSEd, DPhil$^{3}$; Joao H. Bettencourt-Silva, MSc, PhD$^{1}$}

\institutes{
    $^1$IBM Research Europe, Dublin, Ireland;
    $^2$Monash University, Melbourne, Australia;\\
    $^3$Nuffield Department of Population Health, University of Oxford, Oxford, UK\\
}

\maketitle

\noindent{\bf Abstract}

\textit{% !TEX root = main.tex
The wide adoption of Electronic Health Records (EHR) has resulted in large amounts of clinical data becoming available, which promises to support service delivery and advance clinical and informatics research. 
Deep learning techniques have demonstrated performance in predictive analytic tasks using EHRs yet they typically lack model result transparency or explainability functionalities and require cumbersome pre-processing tasks.
Moreover, EHRs contain heterogeneous and multi-modal data points such as text, numbers and time series which further hinder visualisation and interpretability.
This paper proposes a deep learning framework to: 1) encode patient pathways from EHRs into images, 2) highlight important events within pathway images, and 3) enable more complex predictions with additional intelligibility. The proposed method relies on a deep attention mechanism for visualisation of the predictions and allows predicting multiple sequential outcomes.
%healthcare data representation by transforming EHRs into images that can be easily visualized and a deep learning method that has ability to highlight important events for explainable predictions.
%
%In this paper we demonstrate the performance of the proposed method using MIMIC-III.
% TODO: Joao: need to revise last sentence; still have 20words in the 150 word limit.

}

\section{Introduction}
\noindent
% !TEX root = main.tex

Electronic health records (EHR) are essential in supporting healthcare practitioners track their patients and deliver services. Details from patients' encounters are routinely collected in EHRs and, despite advances in machine learning and statistical modeling, these data are heterogeneous and difficult to reuse. The expectation is that routinely collected data together with deep learning techniques may help drive personalised medicine and improve the quality of healthcare service delivery. Typical analytic tasks include disease classification or prediction of clinical events and a number of different deep learning architectures has been identified\cite{xiao_pmid29893864}. 

%Feature engineering continues to be a bottleneck when creating predictive systems from EHRs and concept embeddings have been used to algorithmically deriving feature representations of clinical concepts or phenotypes from EHR data \cite{xiao_pmid29893864}. 

Most analytic architectures rely on EHR data which has undergone substantial pre-processing steps to transform or model these data before it is later ingested for analysis.

EHR data has been previously modeled as pathways\cite{jhbsPathways} describing patient trajectories through time, and studies have also combined multi-modal information (e.g. static demographic variables with patient vitals and text notes) for prediction tasks with some degree of explainability \cite{pmlr-v68-suresh17a}. Other studies have represented EHR information as a matrix of ICD9 codes versus number of visits, and used convolutional neural networks (CNNs) for prediction\cite{pmid29994534}.

The way in which data is modeled and represented can help support both the prediction of clinical events and the interpretability of the results, the latter being particularly important in healthcare. Bringing models into real-world use requires understanding the mechanisms by which models operate and this level of transparency is currently challenging to achieve\cite{xiao_pmid29893864}. 
Attention mechanisms are a common way to provide visual explanations for natural images in neural networks by highlighting the most relevant events in terms of contributions to model prediction or classification.
%
%Specifically, given an image as input, deep attention modules may highlight the important events for the decision of the model and suppresses the unimportant events.
%
Attention mechanisms can be divided into \textit{soft} attention and \textit{hard} attention and their associated weights represent the degree in which the model is paying attention to certain regions of an image. While \textit{soft} attention uses differentiable functions to produce attention weights, \textit{hard} attention involves non-differentiable functions to generates binary weights.
For example, Xu et al. exploited \textit{soft} attention for natural image captioning ~\cite{xu2015show} and several improvements have been made to \textit{hard} attention for the model to be differentiable (e.g. REINFORCE\cite{williams1992simple}).
%%

%
%The ``soft'' attention\cite{bahdanau2014neural} has been firstly introduced in neural machine translation\cite{sutskever2014sequence} (NMT) that significantly improve model's performance as well as in natural imaging domain (e.g., image captioning \cite{xu2015show},  object proposals \cite{jetley2018learn}).
%The attention mechanism is inspired by natural human behavior to focus on specific information in order to archive tasks.
%The ``soft'' represents for the highlighted regions in pathway image are defined by differential functions (e.g., neural network) while other kind of attention is ``hard'' attention which is non-differential by discretely selecting regions such as REINFORCE\cite{williams1992simple} and ``hard'' NMT \cite{luong2015effective}.

% Long short-term memory (LSTM) network  \cite{hochreiter1997long} is a special kind of recurrent neural network (RNN) which has memmory states such hidden state and cell state in order to predict a class in sequence data at each step $i$.
% %
%\natasha{Joao please use following information in intro (not sure where though). It should hopefully justify talks about CNN, RNN and attantion}
In this paper we propose to represent EHR data as an image-like 2D matrix in order to predict a sequence of clinical events while providing some level of intepretation of the results. We extract features from EHRs using CNN techniques and predict sequences of clinical events using a Recurrent Neural Network (RNN) and attention techniques.
RNNs are a well-known method for predicting sequential data. However, they do not perform well in practice, especially because of the vanishing gradient and long-term dependency problems \cite{bengio1994learning}.
Long short-term memory (LSTM) \cite{hochreiter1997long} is a special kind of recurrent neural network that aims to overcome these drawbacks. LSTM consists of different gates (such as input, output and forget gate) that allow it to learn to preserve important information and discard other information in order to effectively predict long sequences. 
However, vanilla LSTM does not consider the previously predicted output from a previous state in order to improve the prediction of current state. A teacher enforcing technique may be used to improve performance by keeping constrains between outputs (i.e. passing the previous prediction to the current state). LSTMs using teacher enforcing have been successfully applied to images and used in natural language processing such as image captioning~\cite{xu2015show}, as well as in sequence to sequence problems \cite{sutskever2014sequence}. 

Previous work encoding patient EHR information using autoencoders, RNNs and attention have been applied to improve model accuracy but have not relied on a combination of 2D representation while exposing attention as a mechanism to visualize important areas within a patient's EHR \cite{new_rajkomar2018scalable, new_choi2016retain, new_suo2018deep, pmid29994534}.

%\thanh{ compared to in EHR applications without teacher enforcing}.

%Interpretation model is strongly required in machine learning model especially in medical domain. A common way is to provide visual interpretability using attention mechanism to highlight the most relevant regions in terms of contributions to model prediction/classification because the attention mechanism mimics human behavior by paying attention the most informative to archive the task. Specifically, given an image as input, deep attention module highlights the important events for the decision of the model and suppresses the unimportant events. 
%Attention mechanism can be divided into ``soft'' attention\cite{bahdanau2014neural} and ``hard'' attention\cite{luong2015effective} categories.  While the “``soft'' one is using differentiable functions to produce attention weights, the ``hard'' involves non-differentiable functions to generates binary weights. Therefore, it has been applied for neural machine translation\cite{bahdanau2014neural} and image visual explanation\cite{jetley2018learn}. \thanh{compared to EHR blabla}

%\thanh{I would propose the skeleton is}
%\begin{itemize}
%    \item Introduce problems that our framework want to solve
%    \item Preivous solution for EHR presentation compared to our pathway. 
%   \item Other LSTM method without teacher enforcing vs our in EHR and predict sequence.
%    \item Intepretation model in EHR others vs our method.
%    \item Summarize all contributions.
%\end{itemize}

This paper proposes \textit{Deep EHR Spotlight}, a framework for predicting and highlighting important clinical events from pathways based on electronic health records. In particular, this paper introduces:
\begin{itemize}
    \item a 2D pathway representation which can be used with two dimensional CNN techniques to improve visual interpretation
    \item a novel deep spotlight model which can highlight important events to help interpret the model's predictions, which may include sequential events, using attention, LSTM with teacher enforcing.
\end{itemize}

Experiments were conducted on a real world dataset MIMIC-III\cite{pmid27219127} and an evaluation was carried out based on performance metrics and a domain expert review.

\section{	Methods}
\noindent
% !TEX root = main.tex
%
% !TEX root = main.tex

\begin{figure}[h]
	\centering
	\includegraphics[width=1\linewidth]{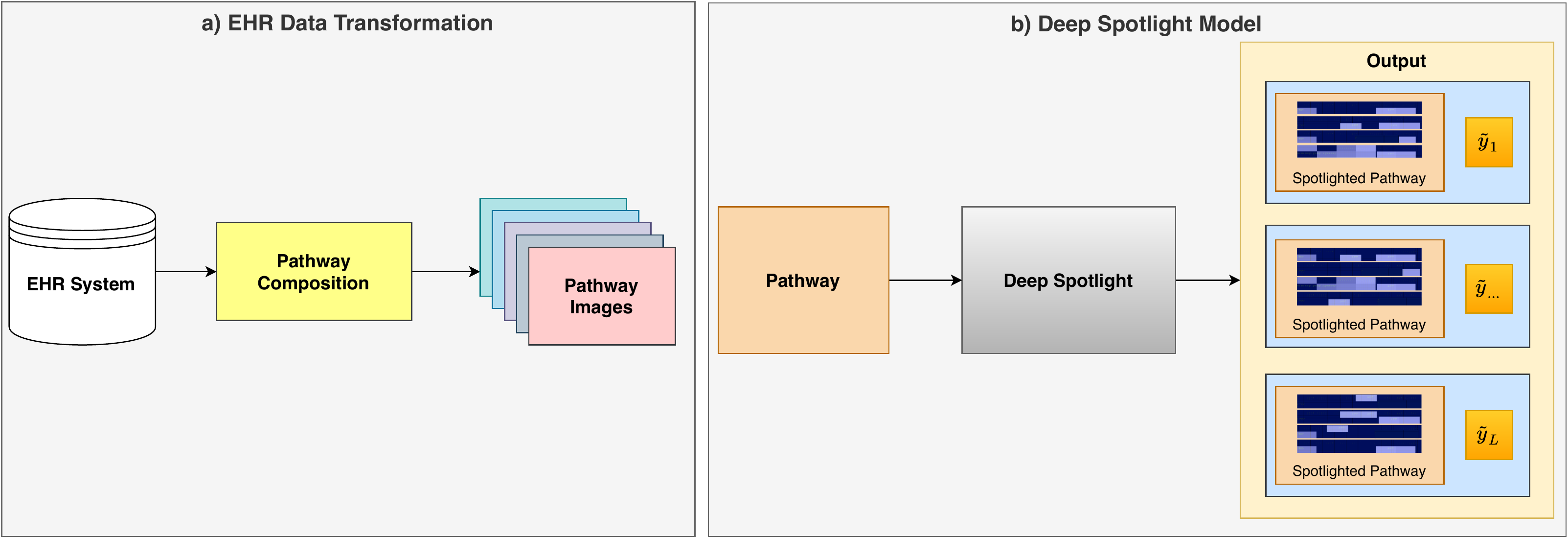}
	\caption{Overview of the framework and its two parts: a) EHR Data Transformation used to compose pathways and b) Deep Spotlight Model used to predicting a sequence of events $y$ and highlight areas that explain predictions $\tilde{y_i}$.}\label{fig:pathway}
\end{figure}

This section describes the proposed framework in detail over two parts as illustrated in Fig.~\ref{fig:pathway}: a) EHR Data Transformation and b) the Deep Spotlight Model.
The first part is an EHR data transformation module which transforms the heterogeneous raw data from an EHR dataset into image-like representations named pathway images $x$ with associated height and width denoted by $h \times w$, respectively.
%
% for enabling spatial correlation for feature extraction using two dimensional CNN technique and better visualization in clinical practice.
%
We also propose a Deep Spotlight Model, described in detail in section 2.2, which uses an attention mechanism that takes as input a pathway, $x$, to predict a sequence of targets (e.g., a sequence of conditions or diseases) $\tilde{y}=\{\tilde{y}_1, \tilde{y}_2, ...,\tilde{y}_L\}, \tilde{y}_i  \in  \mathbb{R}^K,$ where $K$ is the number of possible events to classify at $\tilde{y}_i$ and $L$ is the maximum length of the sequence). The attention mask produced can highlight ('spotlight') the important events that significantly contribute to the model's decision using an attention mechanism described in section \ref{sec:attention}.
%\begin{equation}
%\tilde{y}=\{\tilde{y}_1, \tilde{y}_2, ...,\tilde{y}_L\}, \tilde{y}_i  \in  \mathbb{R}^K,
%\end{equation}  
%where $K$ is the number of event classes to classify at position $\tilde{y}_i$ and $L$ is the maximum length of sequence.
%
Each prediction $\tilde{y}_i$ corresponds to specific highlighted events in the pathway, described in section \ref{sec:pathway_extractor}.

The Pathway Composition module takes the heterogeneous EHR data to produce pathway images consisting of transformed EHR data in a 2-D image-like representation (defined in Section~\ref{sec:pathway_extractor}).
%
%Moreover, the current deep learning methods cannot process text events directly; therefore, we assign a unique integer for each individual text event in all pathways named pathway event encoding.
%
The pathway image $x$ inputs to the Deep Spotlight model to predict a sequence of targets $\tilde{y}$ that are pushed closer to the ground truth $y = \{ y_1 , y_2, ..., y_L \}, y_i  \in  \mathbb{R}^K$ (i.e. the labelled data used for training) by minimizing cross-entropy loss and its attention masks corresponding to each prediction $\tilde{y}_i$. 
The model and approach are described in detail in the next sections.

\subsection{Pathway Data Representation} \label{sec:pathway_extractor}

%Joao still thinking about this, WIP

%pathway data model/dictionary comprised of
%- dimensions (Observations, Procedures, etc)
%- activities/events (a given code appearing at a certain point in a sequence or in time)
%A pathway image is a sequence of activities 

\begin{figure}
	\centering
	\includegraphics[scale=0.5]{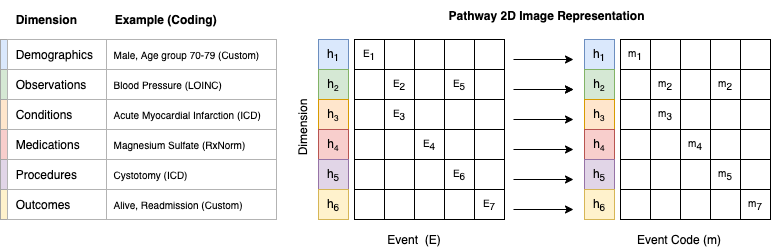}
	\caption{2D-image representation for a given pathway. The y-axis represents the pre-defined dimensions $h$ and the x-axis the pathway events from which codes and values are used.}\label{fig:pathway_2dimage}
\end{figure}

Let $\mathbb{D}$ represent a data model consisting of medical codes, where the $i$-th entry has code $m_i~\in~\mathbb{M}$ where $1 \leq i \leq N$ in a total of $N$ possible codes. Codes may be extracted and mapped directly to terminologies such as ICD or LOINC and each code has an associated dimension, $h_1,...,h_6$, of six pre-defined dimensions that group events of similar nature together (e.g. procedures, observations and medications) as shown in Figure \ref{fig:pathway_2dimage}. 

A pathway describes a patient's hospital admission as a series of events $E=(r, t, m, h)$ where:
\begin{itemize}
    \item $r$ is the patient identifier.
    \item $t$ is the sequence in which events occurred, which can be calculated using the time in days since the day of primary diagnosis for an admission recorded for patient $r$. Events without an associated time have $t=0$.
    \item $m \in \mathbb{M}$ is an event code, such as a specific ICD9, LOINC code, or other, as described in Figure \ref{fig:pathway_2dimage}.
    %\item $v$ is a value associated with medical code $m$, if any. For example the result of a blood test.
    \item $h$ is the dimension associated with medical code $m$ as described in Figure \ref{fig:pathway_2dimage}.
\end{itemize}
Thus, a pathway for a patient $r$ is defined as an ordered set of events $\mathbb{P}=\{E_1,E_2,...,E_M \}$ where:

\begin{enumerate} 
    \item $E_i$ is of the form $(r, t_i, m_i, h_i)$ for $1 \leq i\leq M$,
    \item $t_i \leq t_i+1$ for $1 \leq i \leq M$,
   	\item $M$ is total number of events in a pathway.
\end{enumerate}

Subsequently, a pathway $x$ can be described as a 2-D image by arranging dimensions $h$ across the y-axis and events $E$ across the x-axis as illustrated in Figure \ref{fig:pathway_2dimage}. For each event in the pathway image $x$, the code $m_i$ may be concatenated, resulting in a single value equivalent to each pixel in an image. This representation enables a spatial correlation for feature extraction using two dimensional CNN techniques (e.g., medications and procedures in Fig.~\ref{fig:pathway_2dimage}) and may provide better visualization in practice. The above pathway model is inspired by previous work \cite{jhbsPathways} and may be remodelled to include, for example, the values associated with each event code $m$.

% !TEX root = main.tex
%
%
\subsection{Deep Spotlight Model}
% !TEX root = main.tex
%
\begin{figure}
	\centering
	\includegraphics[width=0.8\linewidth]{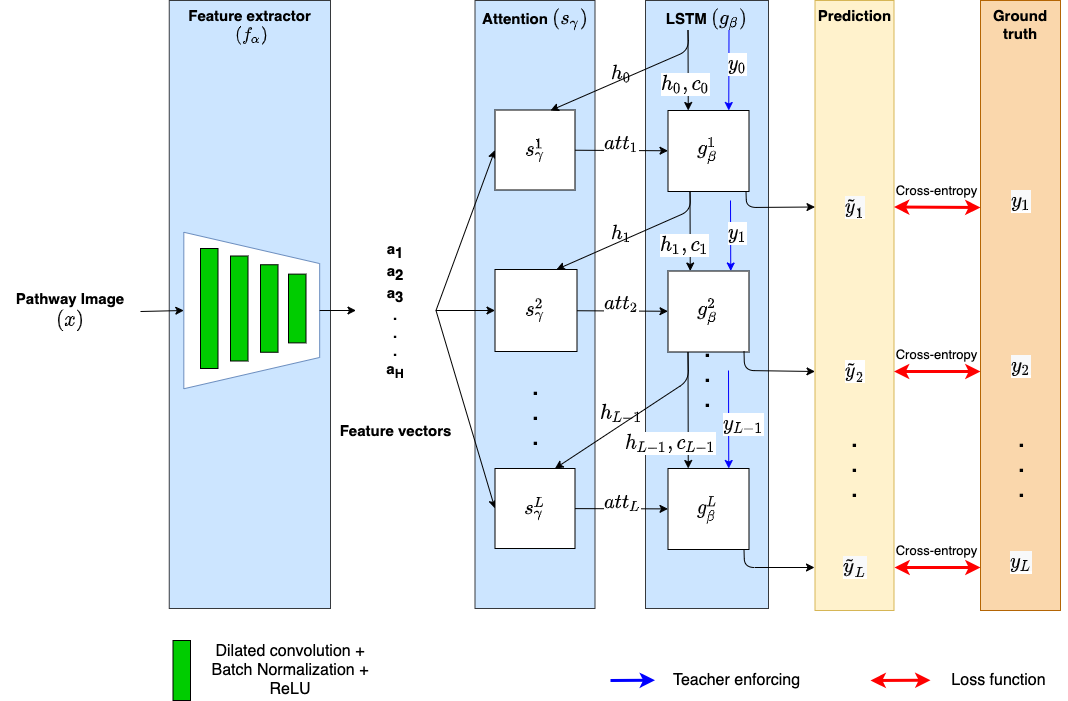}
	\caption{Training process of proposed deep spotlight model consisting of encoder $f$, attention module $s$ and LSTM $g$ parameterized by $\alpha$, $\gamma$ and $\beta$ respectively. The prediction $\tilde{y}$ is pushed close to ground truth $y$ by minimizing cross-entropy loss and current prediction $\tilde{y_i}$ is also enforced during training process by label $y_{i-1}$ called teacher enforcing\cite{williams1989learning}.}
	%\natasha{The Figure must be changed to clearly distinguish functional blocks and data. Also it has to split decoder into 2 parts and open up Encoder. There is no need to have red arrows between prediction and labels. BTW I think we should consider using labels or training data instead of ground truth. Another note in the description of Decoder it says that block g takes both y and $\tilde{y}$ and here there is only y}
    %\natasha{Can we rename this figure to something like "Deep Spotlight Framework Architecture" rather than "Training process"} \thanh{No natasha because training is different from inference}
	\label{fig:model}
\end{figure}
\IncMargin{2em}
\begin{algorithm}[tb!]
	\SetKwData{Left}{left}\SetKwData{This}{this}\SetKwData{Up}{up}
	\SetKwFunction{Union}{Union}\SetKwFunction{FindCompress}{FindCompress}
	\SetKwInOut{Input}{Input}\SetKwInOut{Output}{Output}\SetKwInOut{Initialize}{Initialize}
	\Input{$x$  a single pathway image}
	\Output{$\tilde{y}$ probabilities of predicted conditions\\ $p$ attention masks of predicted conditions }
	%
	%\Initialize{$f_\alpha \leftarrow$ dilated convolution encoder parameterized by $\alpha$\\ 
	%			$g_\beta \leftarrow$ LSTM decoder parameterized by $\beta$\\
	%			$s_\gamma \leftarrow$ determistic soft attention parameterized by $\gamma$}
	%
	\BlankLine
	$a \leftarrow f_\alpha(x)$ \tcp*[f]{$a$:feature vectors extracted by dilated convolution nets}\\
	$h_0 \leftarrow 0$ \tcp*[f]{$h_0$:initial hidden state of LSTM}\\
	$c_0 \leftarrow 0$ \tcp*[f]{$c_0$:initial cell state of LSTM}\\
	$\tilde{y}_0 \leftarrow 0$ \tcp*[f]{$\tilde{y}_0$:initial output at step 0 used as teacher forcing}\\
	$i \leftarrow 1$ \tcp*[f]{$i$:step}\\
	\Repeat{$i\geq L$ or $terminate\_condition$}{
		$p_i, att_i \leftarrow s^i_\gamma(a, h_{i-1})$  \tcp*[f]{$att_i$:attention features for predicting condition $i$}\\
		$\tilde{y}_i, h_i, c_i \leftarrow g^i_\beta(att_i, \tilde{y}_{i-1}, h_{i-1}, c_{i-1})$    \tcp*[f]{$\tilde{y}_i$:predicted condition $i$}\\
		$\tilde{y} \leftarrow y.append(\tilde{y}_i)$  \tcp*[f]{save predicted condition $i$}\\
		$p \leftarrow  p.append(p_i)$  \tcp*[f]{save attention mask of predicted condition $i$}\\
		$h_{i-1} \leftarrow h_i$  \\
		$c_{i-1} \leftarrow c_i$  \\
		$\tilde{y}_{i-1} \leftarrow \tilde{y}_i$  \\
		$i \leftarrow  i + 1$ \\
	}
	\Return $\tilde{y},p$
	\caption{Predicting process of a sequence of medical events $\tilde{y}$  and attention masks $p$ corresponding to each $\tilde{y}_i$ given an input pathway image $x$.}
	\label{alg:onverview}
\end{algorithm}\DecMargin{1em}
Deep Spotlight model is designed to predict a sequence of output events and to highlight input events which contribute to the prediction. We denote the three main modules: feature extractor $f$, attention module  $s$ and long short-term memory module $g$ which are parameterized by $\alpha$, $\gamma$ and $\beta$ respectively.
The feature extractor uses convolutional neural networks $f_{\alpha}$ (CNN) to extract features $a$ from input pathway image. 
The attention module $s_{\gamma}$ produces attention masks $p$ with values from [0; 1] in order to highlight important events that significantly contributed to the predictions. This is done by increasing the weights for important feature regions of $a$ and decreasing for unimportant ones. The LSTM module is used to generate the predicted sequence targets $\tilde{y}$ by taking the output of the attention module, as shown in Fig~\ref{fig:model}.

Note that we take the dimension (row) from the pathway image which contains a sequence of events that represents our training ground-truth $y$. Therefore, the input image size is $h-1 \times w$. For example, in oder to predict conditions, the condition dimension in Fig.~\ref{fig:pathway_2dimage} may be taken to become ground-truth $y$.

\subsubsection{Feature extractor using dilated convolution ($f_\alpha$)}
\begin{figure}
	\centering
	\includegraphics[width=0.6\linewidth]{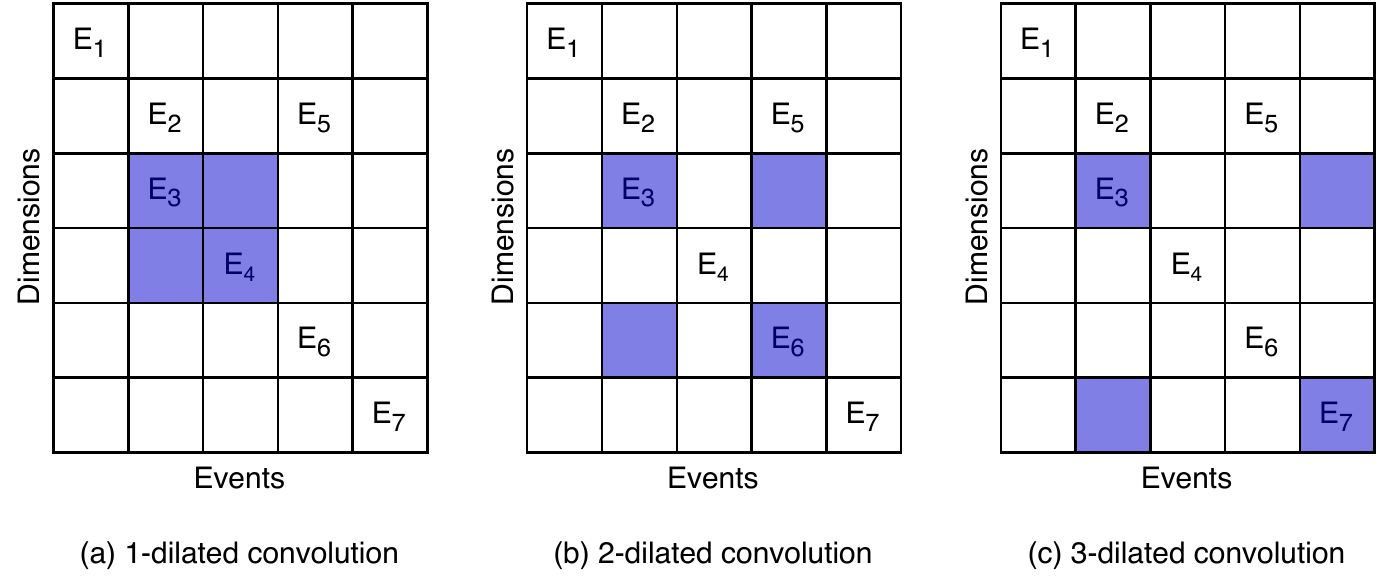}
	\caption{Receptive field expansion of a 2 $\times 2$ convolution filter at different dilation factors on pathway image.}\label{fig:dilatedconv}
\end{figure}

 Vanilla convolutional layers \cite{lecun1998gradient} do not perform well in capturing the global context from images because small blocks of pixels can only be influenced by their filter size\cite{yu2015multi}, especially in sparse signals; however, using larger filter sizes requires more learnable parameters which lead to computational expensive and the \textit{data hungry} problem. 
 Specifically, we observed that pathways extracted from MIMIC-III are highly sparse as empty elements can be 14$\times$ more frequent than encoded events.
 Therefore, to overcome these drawbacks our feature extractor $f_\alpha$ efficiently extracts features from the input pathway image using dilated convolution layers\cite{yu2015multi}.
The motivation for feature extractor architecture is based on the fact that the dilated convolutions support exponentially expanding receptive fields, which is the implicit area captured on the input, while the number of parameters associated with the layer are identical.
In simple intuition in Fig.~\ref{fig:dilatedconv}, the dilated convolution layer is just a convolution layer applied to input with defined gaps.
Moreover, various well-known deep learning architectures use batch normalization\cite{ioffe2015batch} to tackle the internal covariate shift problem in deep neural networks, and ReLU\cite{nair2010rectified} to avoid the vanish gradient problem during training (e.g. ResNet\cite{he2016deep}, DenseNet\cite{huang2017densely}). Thus, our feature extractor consists of layers where each contains a dilated convolution layer, batch normalization and ReLU, as shown in Fig.~\ref{fig:model}. The output from the dilated convolution networks is a set of feature vectors $a$:
 \begin{equation}
a = \{ a_1 , a_2, ..., a_H \}, a_i  \in  \mathbb{R}^F,
\end{equation}
where $H=h' \times w'$ is the number of flatten output feature vectors from the last convolutional layer which corresponds to spatial locations $h-1 \times w$ of the input pathway and $F$ is the number of convolutional filters at the last layer of $f_\alpha$.
%
%\natasha{As a result Encoder produces feature vectors capturing latent dependencies between pathways events occurred chronologically}
By using dilated CNN, we can leverage the pathway image representation by efficiently encoding spatial correlations to feature vectors $a$ which is the input to the attention module (see Section~\ref{sec:attention}). 
%
%\natasha{Highlighting events come very sudden here. There is no explanation on what mechanism was used or what it is based on.}
%This give our approach an ability to selectively focus on highlighted events of input image by selecting a relevant subset of the feature vectors that mostly contributes to prediction (see Section~\ref{sec:attention}).
%
\subsubsection{Attention mechanism ($s_\gamma$) to highlight events }
\label{sec:attention}
In this paper, we use \textit{soft} attention\cite{bahdanau2014neural} to produce attention mask $p$ in order to highlight important events in the pathway image that significantly contribute to the model's prediction. The attention mechanism mimics human behavior when focusing on the most informative regions to make decisions. It can also be efficiently trained using popular gradient-based methods in current deep learning frameworks.
Our attention module $s_{\gamma}$ leverages fully connected neural networks which take feature vectors $a$ and hidden state $h_{i-1}$ from the LSTM (defined in Section~\ref{sec:decoder}) as input in order to generate $score_i$ as in step $i$ in Eq.~\ref{eq:score}.
\begin{equation}\label{eq:score}
score_{i} = s^i_\gamma(a, h_{i-1}).
\end{equation}
%
%
%\natasha{Please consider avoiding "spotlight" or "spotlighting" mask. Instead present is as "a mask that generates a spotlight"}
The attention mask $p_i$, used for visualizing the highlighted areas in Fig.\ref{fig:pathway}, is then calculated by passing $score_i$ through the softmax function that produces attention weights from [0;1], as shown in  Eq.~\ref{eq:m}.
\begin{equation}\label{eq:m}
p_{i} = softmax(score_{i} ).
\end{equation}
%
%\natasha{There is a formal way of calling it something like pooling across filters or something like this. We may need reference here. Also sentence is required to stress on the importance of this attention features and attention mask explaining the purpose and differences between them} \thanh{it's called Global average pooling but it's used for CAM not in attention.}
Finally, attention features $att_i$ are computed using  element-wised multiplication between $a$ and $p_i$ to select important (spotlighted) regions.
\begin{equation}\label{eq:att}
att_{i} = a \odot p_i.
\end{equation}
Attention mask $p_i$ gives a score for each extracted feature vector $a$ which corresponds to a location in the pathway image. The higher the value, the more important the events that contributed to the model's prediction. 

\subsubsection{Long short-term memory network ($g_\beta$) for sequence prediction}
\label{sec:decoder}
An LSTM network is used in our framework to improve the performance of predicting sequences and has memory states, hidden state $h_i$, and cell state $c_i$ at step $i$.
Specifically, during training, the LSTM module takes the most informative extracted features (weighted attention features) $att_i$ from the \textit{soft} attention module's output, previous state ($h_{i-1}$ and $c_{i-1}$) and  previous ground-truth $y_{i-1}$ to predict a target (e.g., a condition or disease) $\tilde{y}_i$ at step $i$ and output information for next prediction ($h_{i}$ and $c_{i}$). Moreover, using  previous ground-truth as an input, we can improve the constrains of the LSTM with the teacher enforcing technique\cite{williams1989learning}, as shown in Fig.~\ref{fig:model}.
% in Eq.~\ref{eq:lstm}.
%%
%\begin{equation}\label{eq:lstm}
%\tilde{y}_i, h_i, cell_i \leftarrow g^i_\beta(att_i, \tilde{y}_{i-1}, h_{i-1}, cell_{i-1}).
%\end{equation}
%
The LSTM predicts elements one by one $\tilde{y}_i$ in $\tilde{y}$ at every step $i$ until reaching terminate conditions.
%
%\natasha{I would just say: The training is reinforced using using Teacher Forcing that .... and explain why}
Note that we do not have previous ground-truth $y_{i-1}$ during the prediction state; therefore, a new prediction $\tilde{y}_i$ is enforced by previous prediction $\tilde{y}_{i-1}$ that improves the predicted sequence's accuracy, as shown in Alg.~\ref{alg:onverview}.
%
%\natasha{This sentence should either be deleted or explained}Moreover, we can give interpretation for every prediction $\tilde{y}_i$ by using corresponding $p_i$.
%

\section{	Results and Discussion}
\noindent
% !TEX root = main.tex

%We'll add some of the limitations here:
%\begin{itemize}
%	\item About the mask
%	\item About the dataset
%	\item About the data model/representation
%\end{itemize}

%\begin{itemize}
    
%    \item Accuracy
%    \item Explainable - show example images and explain their 'stories'. Let's think about ways to evaluate attention (could we propose some ideas for evaluation?). Attention location alone will not tell us if the contents of that area are similar because our content isn't static. Location will only really tell us at what point in the pathway the relevant area is predictive of disease (or outcomes). 
%    \item Stats: how many attended events over overall/attended area: can we add this to  Table 2  summarising the results across all pathways? I can then have a look at the pathway images and maybe if it's easy we can do a rough evaluation with a clinician (asking human to inspect if highlighted events are good)
%    \item Stats of the pathway MIMIC dataset first
%    \item Showing images
%\end{itemize}

This section first describes the results of the transformations to compose pathway images using the MIMIC-III dataset. The performance results of the proposed Deep EHR Spotlight framework are described in section 3.2 and an evaluation with a domain expert is described in section 3.3.

\subsection{Creating Pathway Images from the MIMIC-III Dataset}
Pathway images were produced using the MIMIC-III dataset and based on the definitions provided in \ref{sec:pathway_extractor}. ICD9 codes were reclassified into a smaller number of disease groups of similar codes based on a re-classification system by \textit{Rassekh et. al.}\cite{ICDcodes_pmid21234317}. Other approaches may be used to group ICD codes together for specific use cases, however, this was considered sufficient to train the models presented in this paper and to support building the proposed framework. Selecting a large enough amount of training data was also constrained on the length of the pathways (i.e. the x-axis where event codes are displayed). Overall 58,976 pathway images were produced across 102 different diagnosis groups and 56\% of those had a length of 400 or under. The selection of diagnosis groups was based on balancing the most frequent conditions (diagnosis groups) with the lengths of the pathways in order to provide a large enough training set.
%Table \ref{table:data} shows the total number of pathways for each diagnosis group (condition).
%
Furthermore, the pathways selected also have a sequence of conditions $y$ (based on the diagnosis codes groups) which include at most $L=2$ conditions.
Specifically, sequence $y$ begins with one of the three selected main conditions \{\textit{Birth Outcome, Cerebrovascular Disease, Ischemic Heart Disease}\}, and is followed by an optional second condition from the 99 remaining most frequent conditions. The second condition was also selected based on frequency, as shown in Table~\ref{table:data}.
%
% \natasha{As a suggestion we could change text in the first 3 lines of the table to reflect any secondary condition, e.g. Birth Outcome -> Secondary Condition or we could put sub header or split table into 2}
A total of 11,400 pathway images (i.e. the first three rows in Table~\ref{table:data}) were selected and split 80\% for training and 20\% for testing. The task undertaken to demonstrate the developed framework involves predicting a first main condition in a pathway followed by a second condition. For this reason the pathway image dimension $h$ associated with conditions was removed from the training set. 

\subsection{Performance Evaluation}
The proposed framework was first evaluated by computing peformance metrics: precision, recall and F1 score, as shown in Table~\ref{table:data}. Whilst these metrics are computed by comparing predicted $\tilde{y}_0$ and labeled $y_0$, another metric (Intersection over union (IoU), described later) can be calculated to evaluate the predicted sequence $\tilde{y}$ against a ground-truth (labelled) sequence $y$.
With respect to precision and recall, the proposed method achieved adequate scores (over 90\% F1 scores) for predicting the main condition alone. However, due to the small amount of data and their heterogeneity, F1 scores for the sequence of two conditions were significantly lower. For example, as seen in Table \ref{table:data}, there are very few training samples for \textit{Cerebrovascular Disease $\to$ Arrhythmia} (133 pathways, F1 27\%) and Ischemic Heart Disease $\to$ Hypertension (312 pathways, F1 31\%). It is expected that the proposed method may reach better performance scores for predicting sequences of conditions with sufficiently larger amounts of training data. For example, \textit{Birth Outcome $\to$ Perinatal Condition} (3931 pathways) shows an F1 score of 85\%.
Moreover, a confusion matrix was calculated and is shown in Table~\ref{table:confusion1}. The proposed approach shows good performance on \textit{Birth Outcome, Cerebrovascular and Ischemic Heart Disease} reaching precisions 99.86\%, 91.54\% and 93.31\%, respectively.
We also observed that the performance gradually drops for \textit{Ischemic Heart Disease $\to$ Cardiomyopathy} and \textit{Ischemic Heart Disease $\to$ Hypertension} due to a highly imbalanced dataset.

% Results: Percent (N)
% Condition	Yes, Specific	Yes, General	No
%Cerebrovascular only	0.3 (6)	0.7 (14)	0 (0)
%Cerebrovascular-Arryhthmia	0.75 (15)	0.2 (4)	0.05 (1)
%Cerebrovascular-NeurologDis	0.8 (16)	0.2 (4)	0 (0)
%BirthOutcome only	0.7 (14)	0.3 (6)	0 (0)
%BirthOutcome - Surgery	0.85 (17)	0.15 (3)	0 (0)
%BirthOutcome - Perinatal Condition	0.9 (18)	0.1 (2)	0 (0)
%Ischemic heart disease - only	0.75 (15)	0.25 (5)	0 (0)
%Ischemic heart disease - cardiomyopathy	0.35 (7)	0.6 (12)	0.05 (1)
%Ischemic heart disease - hypertension	0.4 (8)	0.6 (12)	0 (0)

% \begin{table}
% 	\begin{tabular}{ |c|c|c|c|c| } 
% 		\hline
% 		& Abbreviation & Train & Test & Total \\
% 		\hline
% 		BirthOutcome &BO & 6016 & 659 & 6675 \\ 
% 		\hline
% 		Cerebrovascular &Ce& 1688 & 205 & 1893 \\
% 		\hline
% 		IschaemicHeart & IH & 2556 & 276 & 2832 \\
% 		\hline
%         \hline
% 		BirthOutcome-NonHelthSurgery & BO-NH & 569 & 47 & 616 \\
% 		\hline
% 		BirthOutcome-PerinatalCondition & BO-PC & 3510 & 421 & 3931\\
% 		\hline
% 		Cerebrovascular-Arryhthmia &Ce-Ar & 122 & 11 & 133 \\
% 		\hline
% 		Cerebrovascular-NeurologDis &Cr-Ne & 272 & 38 & 310\\
% 		\hline
% 		IschaemicHeart-CardiomyopathyHF & IH-Ca & 399 & 51 & 450\\ 
% 		\hline
% 		IschaemicHeart-Hypertension & IH-Hy & 299 & 22 & 321 \\
% 		\hline
% 	\end{tabular}
% 	\caption{MIMIC number of pathways}
% 	\label{table:data}
% \end{table}

\begin{table}
	\centering
	\begin{tabular}{ |c|c|c||c|c|c|c| } 
		\hline
		\textbf{Condition (Diagnosis Group)} & \textbf{Abbreviation} & \textbf{Pathways (N)} & \textbf{Precision} & \textbf{Recall} & \textbf{F1}  \\
		\hline \hline
		Birth Outcome &BO & 6675 & 0.999 & 0.996 & 0.997  \\ 
		\hline
		Cerebrovascular Disease &Ce & 1893 & 0.916 & 0.903 & 0.909 \\
		\hline
		Ischemic Heart Disease & IH & 2832 & 0.933 & 0.948 & 0.941 \\
		\hline
        \hline
		BO $\to$ Elective Surgery & BO-NH & 616 & 0.623 & 0.494 & 0.551  \\
		\hline
		BO $\to$ Perinatal Condition & BO-PC & 3931 & 0.912 & 0.797 & 0.850 \\
		\hline
		Ce $\to$ Arrhythmia &Ce-Ar & 133 & 0.529 & 0.187 & 0.277  \\
		\hline
		Ce $\to$ Neurological Disorder &Cr-Ne  & 310 & 0.914 & 0.348 & 0.504 \\
		\hline
		IH  $\to$ Cardiomyopathy \& HF & IH-Ca & 450 & 0.463 & 0.221 & 0.299 \\ 
		\hline
		IH $\to$ Hypertension & IH-Hy & 321 & 0.583 & 0.219 & 0.318 \\
		\hline
	\end{tabular}
	\caption{Performance of the proposed framework (precision, recall and F1) for predicting a given condition or a sequence of two conditions. This table also shows the number of pathway images for each condition. The diagnoses codes used for each condition were selected based on an ICD9 re-classification system \cite{ICDcodes_pmid21234317}.}
	\label{table:data}
\end{table}
\begin{table}
	\centering
	\begin{tabular}{l|c|c|c|c|c|c|c|c|c|c|c|}
		\multicolumn{2}{c}{}&\multicolumn{9}{c}{\textbf{Predicted Condition}}\\
		\cline{3-11}
		\multicolumn{2}{c|}{}& BO & BO-NH & BO-PC & Ce  &  Ce-Ar &  Cr-Ne & IH & IH-Ca & IH-Hy\\
		\cline{2-11}
		& BO & \textbf{99.86} & 0 & 0 & 0 & 0 & 0 & 0.13 & 0 & 0\\
		\cline{2-11}
		& BO-NH & 0 & \textbf{62.32} & 26.08 & 0 & 0 & 0 & 0 & 0 & 0 \\
		\cline{2-11}
		\multirow{2}{*}{\rotatebox[origin=c]{90}{\textbf{True Condition}}}& BO-PC & 0 & 4.58 & \textbf{91.06} & 0 & 0 & 0 & 0 & 0 & 0 \\
		\cline{2-11}
		& Ce & 1.40 & 0 & 0 & \textbf{91.54} & 0 & 0 & 7.041 & 0 & 0 \\
		\cline{2-11}
		& Ce-Ar & 0 & 0 & 0 & 5.88 & \textbf{52.94}& 5.88 & 0 & 0 & 0\\
		\cline{2-11}
		& Cr-Ne & 0 & 0 & 0 & 0 & 5.714 & \textbf{91.42} & 0 & 0 & 0 \\
		\cline{2-11}
		& IH & 0 & 0 & 0 & 6.68 & 0 & 0 & \textbf{93.31} & 0 & 0 \\
		\cline{2-11}
		&  IH-Ca & 0 & 0 & 0 & 0 & 0 & 0 & 4.87 & \textbf{46.341} & 14.63 \\
		\cline{2-11}
		& IH-Hy & 0 & 0 & 0 & 0 & 0 & 0 & 0 & 14.70 &\textbf{61.76} \\
		\cline{2-11}
	\end{tabular}
	\caption{Confusion matrix whose vertical  axis shows ground-truth conditions and  horizontal axis illustrates predicted conditions (\%).}
	\label{table:confusion1}
\end{table}
%
%
%The peformance of framework is also evaluated by using Bilingual Evaluation Understudy ($BLEU$) score. \joao{bring in sentence explaining BLEU scores and citation, what is BLEU score giving us? compare to compare with ground truth} We compute $BLEU_1$ and $BLEU_2$ that reached 0.749 and 0.53, respectively. 

Intersection over union (\textit{IoU}) is a metric for evaluating the differences between predicted sequence $\tilde{y}$ and ground-truth sequence $y$.
\textit{IoU} is measured by overlapping  $\tilde{y}$ and $y$ as shown in Eq.~\ref{eq:iou}

\begin{equation}
\label{eq:iou}
IoU = 2 \dfrac{\tilde{y} \cap y}{\tilde{y} \cup y},
\end{equation} 
where $IoU$ is equal to 1  if two sequences are perfectly matched. The calculated average $IoU$ metric across all pathways was 0.75, showing agreement between the predicted sequence against the ground-truth.

\begin{table}
	\centering
	\begin{tabular}{ |c|c|c|c| } 
		\hline
		\textbf{Predicted Condition} & \textbf{Specifically Related} &\textbf{Related} & \textbf{Not Related} \\
		\hline \hline
		Birth Outcome (BO)  & 14 & 6 & 0\\ 
		\hline
		Cerebrovascular Disease (Ce) & 6 & 14 & 0 \\
		\hline
		Ischemic Heart Disease (IH)  & 15 & 5 & 0 \\
		\hline
		\hline
		(BO $\to$) Elective Surgery & 17 & 3 & 0\\
		\hline
		(BO $\to$)  Perinatal Condition & 18 & 2 & 0\\
		\hline
		(Ce $\to$)  Arrhythmia & 15 & 4 & 1\\
		\hline
		(Ce $\to$)  Neurological Disorder & 16 & 4 & 0\\
		\hline
		(IH $\to$)  Cardiomyopathy \& HF  & 7 & 12 & 1\\ 
		\hline
		(IH $\to$) Hypertension & 8 & 12 & 0\\
		\hline
	\end{tabular}
	\caption{Evaluation of the top 20 events highlighted by the attention mask for each predicted condition where $( \cdot )$ shows the first predicted condition.}
	\label{table:doctor}
\end{table}

\section{Domain Expert Evaluation}
\begin{figure}[h]
\centering
\includegraphics[width=1\linewidth]{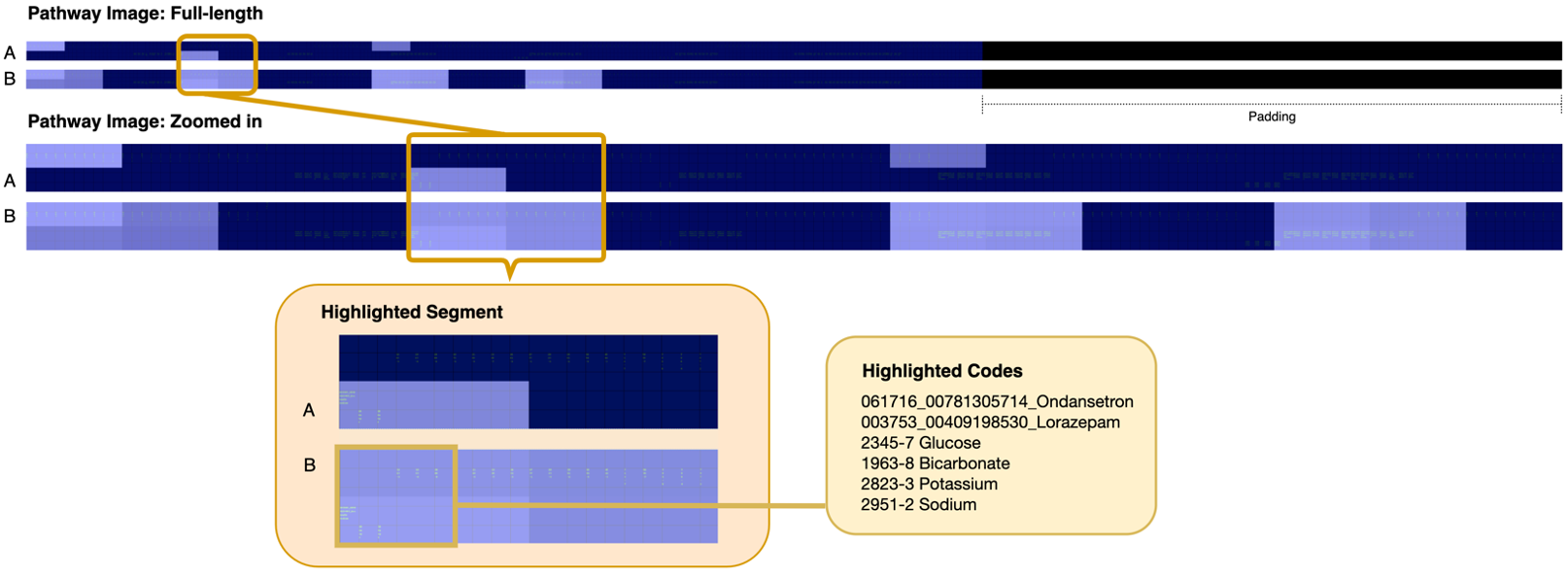}
\caption{Example of a pathway image (output of the proposed framework) showing the highlighted areas that predicted the pathway's conditions: Cerebrovascular disease alone (A) and Cerebrovascular disease followed by a Neurological Disorder (B). The full-length image shows a darker area denoting padding.}
\label{fig:pathway_images_example}
\end{figure}

The framework was further evaluated with respect to the attention mask and whether it is highlighting important events across all pathways.
While the attention mask $p_i$ values range between 0 and 1, a threshold was set at 0.9 to obtain the top 20 events highlighted by the mask in all pathways of the testing set.
Several events may be highlighted in each pathway image, however, only those that contributed most significantly to the predicted condition $\tilde{y}_i$ were selected (i.e., corresponding to the locations of threshold=0.9 in $p_i$). 
With the help of a domain expert, the top 20 events were inspected and a determination was made on whether the event codes were: specifically related to the predicted condition(s), generally related (not necessarily specific to the predicted condition), or not related, as shown in Table~\ref{table:doctor}.
Despite this evaluation being carried out in a small sample of the highlighted events and by a single observer, it provides reassuring results that most events highlighted as important were indeed related to the predicted condition. One of the limiting factors in this evaluation is the absence of values for the event codes. For example, a large proportion of events highlighted are LOINC codes referring to blood tests and data about the results of the blood tests was not included in our experiments. As described in section \ref{sec:pathway_extractor}, adding values to each event code is possible, however, that would introduce additional sparseness and more training data would be required to test the developed framework.
Figure \ref{fig:pathway_images_example} shows an example of a pathway image and the prediction results provided by the proposed framework as highlighted areas. Pathway image A) in Fig.~\ref{fig:pathway_images_example} shows the highlighted areas that are most predictive of \textit{Cerebrovascular} alone, and pathway image B) shows highlighted areas predicting \textit{Cerebrovascular $\to$ Neurological Disorder} for the same pathway. Fig.~\ref{fig:pathway_images_example}  also shows the pathway image when zoom is applied. A selected segment of the highlighted area was further expanded into text for readability and shows an example of the event codes.
%Pathways used in the example figure:
%A: id/[124002 76761 249]--Cerebrovascular(1.000).png
%B: id/[124002 76761 249]--NeurologDis(0.455).png

\section{	Conclusion}
\noindent
This paper proposes a new framework for predicting and highlighting important clinical events from electronic health records (EHRs). In particular, this paper proposes to transform EHR data into pathways and 2D pathway images, which can then be used with two dimensional CNN techniques to support visual interpretation. The proposed Deep EHR Spotlight framework can highlight regions in pathway images which are particularly important for the predictions. In this paper we used the MIMIC-III dataset, which produced highly sparse pathway images. The performance results were adequate (i.e. F1 scores $> 90\%$ ) when a larger number of training data was available. The top events appearing in highlighted masks were also evaluated by a domain expert and found to be mostly related and specific to the predicted conditions. These are reassuring results that demonstrate the value in the proposed approach. However, further work is needed to substantiate these results, improve the methods for a specific use case and compare with other techniques and approaches. As future work we are planning to test the proposed framework on significantly larger datasets as well as remodeling the pathway images and their dimensions for more specific clinical use cases. This will allow predicting individual conditions and taking into account events with associated codes (e.g. haemoglobin test) and respective values (e.g. 20 g/dL). Further work is also needed to continue to evaluate the events highlighted by the attention mask for different thresholds.

\makeatletter
\renewcommand{\@biblabel}[1]{\hfill #1.}
\makeatother

\bibliographystyle{unsrt}
\bibliography{bibfile}

\end{document}